\documentclass[10pt,twocolumn,letterpaper]{article}

\usepackage{cvpr}

\newcommand{\TODO}[1]{\textbf{\color{red}[TODO: #1]}}
\renewcommand{\TODO}[1]{}

\usepackage{microtype}

\usepackage{listings}
\definecolor{cvprblue}{rgb}{0.21,0.49,0.74}
\usepackage[pagebackref,breaklinks,colorlinks,allcolors=cvprblue]{hyperref}

\def\paperID{****}
\def\confName{CVPR}
\def\confYear{2026}

\title{VISTA: Dense Multi-Label Classroom Coding with Vision-Language Models}

\author{%
Andrew Franck\textsuperscript{1}, Brendan Ng\textsuperscript{1}, Ben Fitzgerald\textsuperscript{1}, Zane Derrod\textsuperscript{1}, Chris Cianci\textsuperscript{1}, Chris Craney\textsuperscript{2}\\[6pt]
{\small\textsuperscript{1}Department of Computer Science, Occidental College,\quad
\textsuperscript{2}Department of Chemistry, Occidental College}
}

\begin{document}
\maketitle

\begin{abstract}
Video-language benchmarks are usually constructed by the dataset authors without published reliability statistics, leaving the noise floor of the construct unknown. We argue that multimodal benchmarking benefits from methods taken from research communities that have already invested in strategies to ensure reliability. We illustrate the case with the Classroom Observation Protocol for Undergraduate STEM (COPUS): a 24-code multi-label observation instrument with a decade of peer-reviewed reliability literature. We recast COPUS as a video benchmark for multimodal foundation models, where it provides a dense set of structured labels (a 24-dimensional binary vector every 2 minutes across a 50--90 minute lecture), an externally validated vocabulary, and established literature that provides a per-code reliability target based on human evaluators. Annotations in our evaluation corpus are produced by a 5-person human-evaluator panel whose consensus matrix is our reference. We propose \textsc{VISTA}, a baseline that runs MiniCPM-V-4.5 over a dense sliding window, refines its per-window outputs with a lightweight multi-layer perceptron (MLP) head trained on top of the frozen backbone, and max-pools the resulting predictions onto the 2-minute COPUS grid. On three held-out chemistry lectures, \textsc{VISTA} reaches $80.1\%$ restricted macro accuracy versus $74.9\%$ for the zero-shot variant, with the largest residual errors on visually similar instructor codes (\textsc{RtW}) and on rare audio-dependent codes. We characterise three systematic failure modes (audio-partial observability, fine-grained group-work discrimination, long-tail recall) and release the benchmark tooling, prompts and baseline code at \url{https://github.com/ajfranck/VISTA}.
\end{abstract}

\section{Introduction}
\label{sec:introduction}

Benchmark renewal in video-language modelling is increasingly a data problem.
Video-language benchmarks such as Video-MME~\cite{videomme},
MVBench~\cite{mvbench}, LongVideoBench~\cite{longvideobench}, and
EgoSchema~\cite{egoschema} have spread quickly, but all share structural limitations.
First, although several of them use multi-stage annotation pipelines, the ground truth
is constructed entirely by the dataset authors and none report formal inter-rater
reliability statistics. The noise floor of the construct is unknown, and there is no externally validated
reliability ceiling against which model error can be compared. Second, the tasks are
dominated by short-clip question--answer pairs; dense labelling at regular intervals
throughout the video is rare.
Third, most benchmarks define their own custom label set, so cross-benchmark scores reflect dataset choices as much as model skill.

We argue that the multimodal community underutilises a different kind of
benchmark: instruments taken from research communities that have spent years
validating their reliability. As a concrete case study, we propose the
Classroom Observation Protocol for Undergraduate STEM (COPUS)~\cite{smith2013copus}
as a video benchmark; COPUS is the standard classroom-observation protocol,
although other observation protocols such as RTOP~\cite{rtop} and TDOP~\cite{tdop}
have also been used.
COPUS partitions a 50--90 minute lecture
into 2-minute intervals and, for each interval, asks a trained observer to mark which
of 24 predefined behaviours (13 student, 11 instructor) occurred. The protocol has
been widely used in the STEM education literature, has a published inter-rater
reliability record, and is the operational target of education researchers
who would benefit from an automated coder. Prior automated analysis of
classroom video and audio~\cite{wang2014audioclass,donnelly2016multisensor} has not, to
our knowledge, used a modern vision-language model to predict the COPUS code set.


Re-interpreted as a vision-language task, COPUS has four properties that current
video-LM benchmarks largely neglect.
It provides dense, long-horizon structure
(a 60 min lecture contains 30 intervals $\times$ 24 codes $=$ 720 
predictions per video); an externally validated
vocabulary standard across institutions and peer-reviewed~\cite{smith2013copus}; a
published inter-rater reliability literature reporting aggregate Cohen's $\kappa$ of
$0.79$--$0.87$ across trained-observer pairs~\cite{smith2013copus}, giving an empirical
target for human-level agreement on the protocol;
and fine-grained, multi-modal distinctions (\eg Clicker-Group vs Worksheet-Group
depends on small objects in the scene) that are only partially observable from vision,
motivating audio fusion and cross-modal alignment work.

We propose COPUS as a multimodal foundation-model benchmark and characterise what
distinguishes it from existing video-LM evaluations; provide \textsc{VISTA} (Vision Instrument for STEM
Teaching Activity), a MiniCPM-V-4.5~\cite{minicpmv} baseline with a structured
multi-label prompt and max-pool aggregator matching the protocol's any-occurrence
semantics; release evaluation tooling and the prompt
vocabulary; and discuss failure modes that COPUS isolates more cleanly
than current video-LM benchmarks.

\section{The COPUS Benchmark}
\label{sec:benchmark}

Let $\mathcal{A}$ denote the 24 COPUS codes (Table~\ref{tab:copus-codes}) and let
$I_k = [120k,\; 120(k{+}1))$ seconds denote the $k$-th 2-minute interval of a lecture of
duration $T$ seconds, $K = \lceil T/120 \rceil$. A system must produce, for each interval
$k$ and each $a \in \mathcal{A}$, a binary label $y_{a,k} \in \{0,1\}$. The reference
matrix $Y \in \{0,1\}^{|\mathcal{A}|\times K}$ is the consensus of five human
evaluators, each independently coding every lecture following the standard COPUS
protocol~\cite{smith2013copus}. Evaluation uses per-code accuracy for non-rare codes,
recall for rare codes (where accuracy is dominated by true negatives), and restricted
macro accuracy averaged over codes that occur in the evaluation set.

\begin{table}[t]
\centering
\caption{The 24 COPUS codes. Each interval receives a binary annotation per code.}
\label{tab:copus-codes}
\footnotesize
\setlength{\tabcolsep}{4pt}
\begin{tabular}{@{}lll@{\hspace{0.8em}}lll@{}}
\toprule
Code & Who & Description & Code & Who & Description \\ \midrule
L   & Stu & Listening          & Lec & Ins & Lecturing \\
Ind & Stu & Individual think.  & RtW & Ins & Real-time writing \\
CG  & Stu & Clicker group      & FUp & Ins & Follow-up \\
WG  & Stu & Worksheet group    & PQ  & Ins & Posing question \\
OG  & Stu & Other group        & CQ  & Ins & Clicker question \\
AnQ & Stu & Answering Q.       & AnQ & Ins & Answering Q. \\
SQ  & Stu & Asks question      & MG  & Ins & Moving/guiding \\
WC  & Stu & Whole-class disc.  & 1o1 & Ins & One-on-one \\
Prd & Stu & Prediction         & D/V & Ins & Demo/video \\
SP  & Stu & Presentation       & Adm & Ins & Administration \\
TQ  & Stu & Test/quiz          & W   & Ins & Waiting \\
W   & Stu & Waiting            & & & \\
O   & Stu & Other              & & & \\ \bottomrule
\end{tabular}
\end{table}

Our evaluation corpus consists of 13 human-annotated university chemistry lectures, recorded with a single fixed-angle
camera covering both instructor and students. Each lecture was independently coded by
five trained human evaluators using the COPUS spreadsheet format, and the resulting
five evaluation matrices were merged into a single consensus matrix that we treat as the ground
truth $Y$. Our comparison tooling reads those matrices directly, so there is no
re-annotation step and no ground-truth transformation. For this paper we report results
on three full lectures whose code-frequency values span the full spectrum of rarity.
We use the published COPUS inter-rater reliability literature as the human noise floor
for our corpus rather than computing Fleiss' $\kappa$ directly on our 5-rater matrices;
the implications of this choice are discussed in \S\ref{sec:discussion}.

Four properties of COPUS surface failure modes that current short-form video-LM
benchmarks do not isolate: multi-label co-occurrence (intervals can contain several
simultaneous codes, yielding higher label density than Charades-style activity
sets~\cite{charades,piergiovanni2019multi}); fine-grained spatial reasoning;
interval--event granularity mismatch, where events much shorter than the 2-minute
interval must still be counted as present (encouraging localisation-style
models~\cite{actionformer}); and audio-partial observability, since question and
discussion codes are defined partly by spoken intent rather than visible behaviour.
Across our corpus, the codes \textsc{Lec}, \textsc{L}, \textsc{RtW} occur in $>$80\% of
intervals while \textsc{Adm}, \textsc{SP}, \textsc{TQ} occur in $<$5\%. These are rare but important
events that are statistically marginal in any naturally collected corpus.

\paragraph{Published reliability as context.}
Smith et al.~\cite{smith2013copus} report aggregate Cohen's $\kappa$ of $0.79$--$0.87$
across trained-observer pairs (averaged over all codes). Per-code reliability is
reported as Jaccard similarity rather than $\kappa$ because constant codes
(always-absent or always-present) break $\kappa$'s chance correction. Several rare
codes (\textsc{SP}, \textsc{Prd}, \textsc{TQ}) did not occur in their validation
corpus, so no published per-code statistic exists for them.

\section{Baseline Pipeline}
\label{sec:method}

\begin{figure*}[t]
    \centering
    \includegraphics[width=0.95\linewidth]{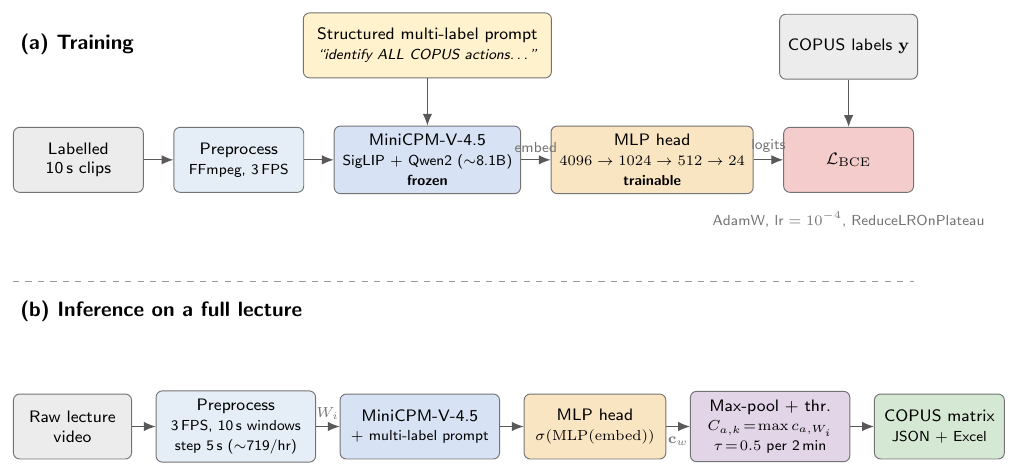}
    \caption{The \textsc{VISTA} baseline. (a)~Training: a 3-layer multi-layer
    perceptron (MLP) head is trained on top of the frozen MiniCPM-V-4.5 backbone with
    binary cross-entropy (BCE) loss against multi-label clip annotations. The
    multi-label prompt is identical at training and inference. (b)~Inference: a full lecture is preprocessed to
    3\,FPS and split into 10\,s windows; per-window
    24-dimensional confidences are max-pooled across windows overlapping each 2-minute
    COPUS interval and thresholded at $\tau = 0.5$.}
    \label{fig:pipeline}
\end{figure*}

We provide a strong baseline (\textsc{VISTA}) whose modular structure makes three design decisions easy to vary independently: temporal granularity, prompt structure, and aggregation operator. The pipeline
(Fig.~\ref{fig:pipeline}) has three stages: 3\,FPS preprocessing, sliding-window
vision-language model (VLM) inference with a multi-label prompt, and max-pool
aggregation onto the 2-minute COPUS grid.

COPUS is defined over 2-minute intervals, but the events that determine a code's presence
(\eg a 7 second \textsc{RtW}) may be much shorter. In the spirit of
sparse temporal segment sampling~\cite{wang2016tsn}, we use a dense sliding window
$W_i = [is, is + \Delta t]$, $\Delta t{=}10$\,s, $s{=}5$\,s, giving 50\% overlap and
719 windows per 60-minute lecture. Any behaviour lasting $\geq$5\,s appears in multiple
VLM predictions and the aggregation step sees redundancy.

One naive approach is to query each of the 24 codes separately, however this costs
24$\times$ more compute, is order-sensitive, and ignores joint structure. We instead
give a single prompt per window instructing the VLM to generate observations for all 24
codes (Listing~\ref{lst:prompt}). On development data, the single-pass prompt produced
more code coverage per window without any observable loss of precision, at 1/24 the inference cost.

\pagebreak
\begin{lstlisting}[caption={Structured multi-label prompt.}, label={lst:prompt}]
Analyze this classroom video and identify ALL COPUS actions that are currently occurring.

COPUS Actions to identify:
{copus_actions_list}

Instructions:
1. Watch the video segment carefully
2. Identify ALL actions that are happening (there can be multiple simultaneous actions)
3. For each action you identify, provide:
- The exact action code from the list above
- Your confidence level (high, medium, or low)
- A brief justification

Format your response as:
DETECTED ACTIONS:
[action_code_1]: [confidence_level] - [brief justification]
[action_code_2]: [confidence_level] - [brief justification]
...
\end{lstlisting}

The VLM's response is converted to a 24-dim confidence vector
$\mathbf{c}_w \in [0,1]^{24}$ either by parsing the ``DETECTED ACTIONS'' block
or by matching wording against a keyword dictionary based on previous model responses.

A 3-layer MLP ($4096 \rightarrow 1024 \rightarrow 512 \rightarrow 24$) is trained on
top of the frozen VLM with BCE loss on a supervised clip corpus drawn from the 10
lectures held out of evaluation, supervised by the same 5-rater consensus. Training uses
AdamW~\cite{adamw}, base learning rate (LR) $10^{-4}$ with
\texttt{ReduceLROnPlateau}, weight decay $10^{-2}$, and gradient clipping at $1.0$.
Because rare codes appear in $<$5\% of intervals, a model trained on a uniform sample
of intervals minimises BCE by predicting negative on every rare class. We address this
by sampling the training clips for diversity: the training set deliberately mixes
intervals with no codes, intervals with a single code, and intervals with multiple
co-occurring codes, so the model has to learn both presence/absence discrimination and
the joint structure of simultaneous behaviours. Full 24-way multi-label BCE
supervision is applied to every clip, so the model is penalised for spurious positives
on any of the 24 codes regardless of which clip type it sees.

Per-window confidences are grouped into the 2-minute COPUS grid by max-pooling over
overlapping windows: $C_{a,k} = \max_{W_i \cap I_k \neq \emptyset} c_{a, W_i}$ and
$\hat{y}_{a,k} = \mathbb{1}[C_{a,k} \geq 0.5]$. The choice of $\max$ over $\mathrm{mean}$
matches the protocol's requirements: COPUS codes a behaviour as present if
it occurs at any point in the interval, so mean pooling would penalise shorter actions/codes.

\section{Results}
\label{sec:experiments}

Inference uses a single NVIDIA L40S. Thus, a 60-minute lecture produces 719 sliding windows
at $\sim$32 GPU-hours, with the dense windowing and 500-token structured-generation
prompt deliberately expensive per window so that brief behaviours are not missed in
long, high-FPS lecture video. We compare two configurations of \textsc{VISTA}: a
fine-tuned (FT) variant with the trained MLP head over MiniCPM-V-4.5 hidden states,
and a zero-shot (ZS) variant using the keyword-parsed VLM output directly. MiniCPM-V-4.5
was chosen over other open-weights VLMs (Qwen2-VL~\cite{qwen2},
VideoLLaMA\,2~\cite{videollama2}, LLaVA-Video~\cite{llavavideo}) because its
SigLIP-style~\cite{siglip} visual encoder runs on-prem on consumer hardware, which is
necessary for our access-restricted classroom video. Per-code accuracy is computed against the five-rater human consensus matrix.
For rare codes (\textsc{SP}, \textsc{Prd}, \textsc{SQ}) we report recall rather than
accuracy: because positive labels are sparse, a model that simply predicts negative
achieves $\geq$95\% accuracy on these codes despite identifying none of the actual
occurrences.

\begin{table}[t]
\centering
\caption{Per-code results across 3 lectures against the 5-rater consensus. Accuracy
is reported for high- and mid-frequency codes, and recall is reported for rare codes.
Restricted macro accuracy is averaged over the
$m{=}$10 COPUS codes with $\geq 1$ positive in the evaluation set.
Codes absent from the corpus are excluded because predicting all-negative on an
absent code trivially yields 100\% accuracy and would inflate the macro.}
\label{tab:results}
\footnotesize
\setlength{\tabcolsep}{6pt}
\renewcommand{\arraystretch}{1.05}
\begin{tabular}{@{}l cc@{}}
\toprule
Code & Ours (FT) & Ours (ZS) \\
\midrule
\multicolumn{3}{@{}l}{\textit{High-frequency (accuracy, \%)}} \\
Lec   & 85.3  & 74.4 \\
L     & 89.7  & 78.5 \\
RtW   & 53.3  & 44.7 \\
MG    & 100.0 & 91.9 \\
\midrule
\multicolumn{3}{@{}l}{\textit{Mid-frequency (accuracy, \%)}} \\
PQ          & 82.0 & 82.0 \\
AnQ (Ins.)  & 76.0 & 74.0 \\
CG / WG     & 78.0 / 72.0 & 79.0 / 75.0 \\
\midrule
\multicolumn{3}{@{}l}{\textit{Rare (recall, \%)}} \\
SP  & 80.0 & 70.0 \\
Prd & 70.0 & 65.0 \\
SQ  & 70.0 & 65.0 \\
\midrule
Restricted macro acc.\ ($m$ codes, $\geq{}1$ pos.) & \textbf{80.1} & \textbf{74.9} \\
\bottomrule
\end{tabular}
\end{table}

The fine-tuned model agrees with the human consensus on 2013 of 2160 binary predictions
across 3 lectures, and restricted macro accuracy (over the $m{=}$10 codes with $\geq{}1$
positive in the evaluation set) is 80.1\%. For context, the published aggregate
Cohen's $\kappa$ for COPUS observer pairs is $0.79$--$0.87$~\cite{smith2013copus},
so our model's residual error reflects model limitations rather than approaching
the human noise floor. Disagreements are largely on visually similar instructor codes
(\textsc{RtW} alone accounts for the largest single-code error and appears similar to
\textsc{Lec} and instructor \textsc{AnQ}) and on rare codes, where the model misses
20--30\% of true occurrences. These are likely due
to limited exposure during pretraining, even with the diversity-balanced training-clip
sampling described in \S\ref{sec:method}. The zero-shot variant is below the
fine-tuned model by roughly 9~pp on the high-frequency codes, is essentially tied on the
mid-frequency codes, and drops 5-10~pp of recall on rare codes.

\paragraph{Three main failure modes.}
Audio-partial codes (\textsc{SQ}, instructor \textsc{AnQ}, \textsc{CQ}): the model
detects visual correlates (\eg hand raise) but cannot distinguish question from
comment. Fine-grained group-work confusion
(\textsc{CG}/\textsc{WG}/\textsc{OG}): the codes differ only in the small object
students hold, so the model fails to distinguish and defaults to the generic
\textsc{OG}---a data-quality (camera placement) failure as much as a model failure.
Long-tail recall (\textsc{SP}, \textsc{Prd}, \textsc{TQ}): the supervised corpus
contains few examples and the VLM has likely seen few classroom ``predictions'' during
its pretraining, yielding the classic data-mixture problem for rare but pedagogically
important events.

\section{Discussion}
\label{sec:discussion}

The agenda for video-language benchmarks is increasingly limited not by data volume but
by the challenge of creating tasks whose predictions have relevance outside
the multimodal community. COPUS demonstrates a complementary idea: benchmarks that have already been validated by a domain community bring with them published reliability, operational definitions, and a downstream use case.

\paragraph{Limitations.}
Several caveats apply. The evaluation corpus is narrow: three chemistry lectures from
one institution with one fixed-angle camera, so generalisation across STEM disciplines
and recording setups is future work. We use the published COPUS reliability
literature~\cite{smith2013copus,stains2018anatomy} as the human noise floor rather than
computing Fleiss' $\kappa$ on our 5-rater panel; because our evaluators followed the
same protocol this is a reasonable proxy, but a direct $\kappa$ on our matrices is the
natural follow-up. The label distribution is a significant data
challenge in its own right: with some codes appearing in less than 5\% of intervals, an unconstrained
training objective can be trivialised by predicting negative everywhere, so recall must be used to evaluate performance. Our diversity-balanced training-clip sampling mitigates this but does not eliminate
it; constructing better-balanced training data is a clear direction. We have not yet compared to a non-VLM baseline such as VideoMAE~\cite{videomae} nor performed extensive audio fusion.
We have institutional access to the classroom video but can not release it publicly. The open-weights backbone~\cite{minicpmv} lets the evaluation run on local hardware, and we release the pipeline, prompts and evaluation tooling but not the raw video. The system should not be used to evaluate individual instructors: published $\kappa$ on several codes is too low for fair single-person decisions. COPUS-style imported benchmarks offer a low-cost path to reliability-calibrated multimodal evaluation, and the same template should apply to other domains  where validated coding schemes already exist.

{
    \small
    \bibliographystyle{ieeenat_fullname}
    \bibliography{main}
}

\end{document}